%% file: paper.tex
\documentclass[]{jingdong}
\usepackage[toc,page,header]{appendix}
\usepackage{algorithm}
\usepackage{algpseudocode}
\usepackage{xspace}
\input{common}
\usepackage{cleveref}
\usepackage{colortbl}

\theoremstyle{plain}

\theoremstyle{definition}

\theoremstyle{remark}
\usepackage{subcaption}

\usepackage[textsize=tiny]{todonotes}
\usepackage{fontawesome}
\newcommand{\method}{\textsc{CoPD}\xspace}
\newcommand{\fullmethod}{Co-Evolving Policy Distillation\xspace}

\title{Co-Evolving Policy Distillation}
{\small

\author[1,2,*]{Naibin Gu}
\author[1,2,*]{Chenxu Yang}
\author[3,*]{Qingyi Si}
\author[1,2]{Chuanyu Qin}
\author[1,2]{Dingyu Yao}
\authorbreak
\author[1,2,\dagger]{Peng Fu}
\author[1,2]{Zheng Lin}
\author[1]{Weiping Wang}
\author[3]{Nan Duan}
\author[3]{Jiaqi Wang}

\affiliation[1]{Institute of Information Engineering, CAS}
\affiliation[2]{School of Cyber Security, UCAS}
\affiliation[3]{JD.COM}

\contribution[*]{Equal contribution.}
\contribution[\dagger]{Corresponding author.}
}

\input{abstract}

\correspondence{\email{gunaibin@iie.ac.cn}; \email{fupeng@iie.ac.cn}; \email{siqingyi.phoebus@jd.com}}

\begin{document}
\maketitle

\begin{figure*}[!ht]
\centering
\includegraphics[width=0.99\textwidth]{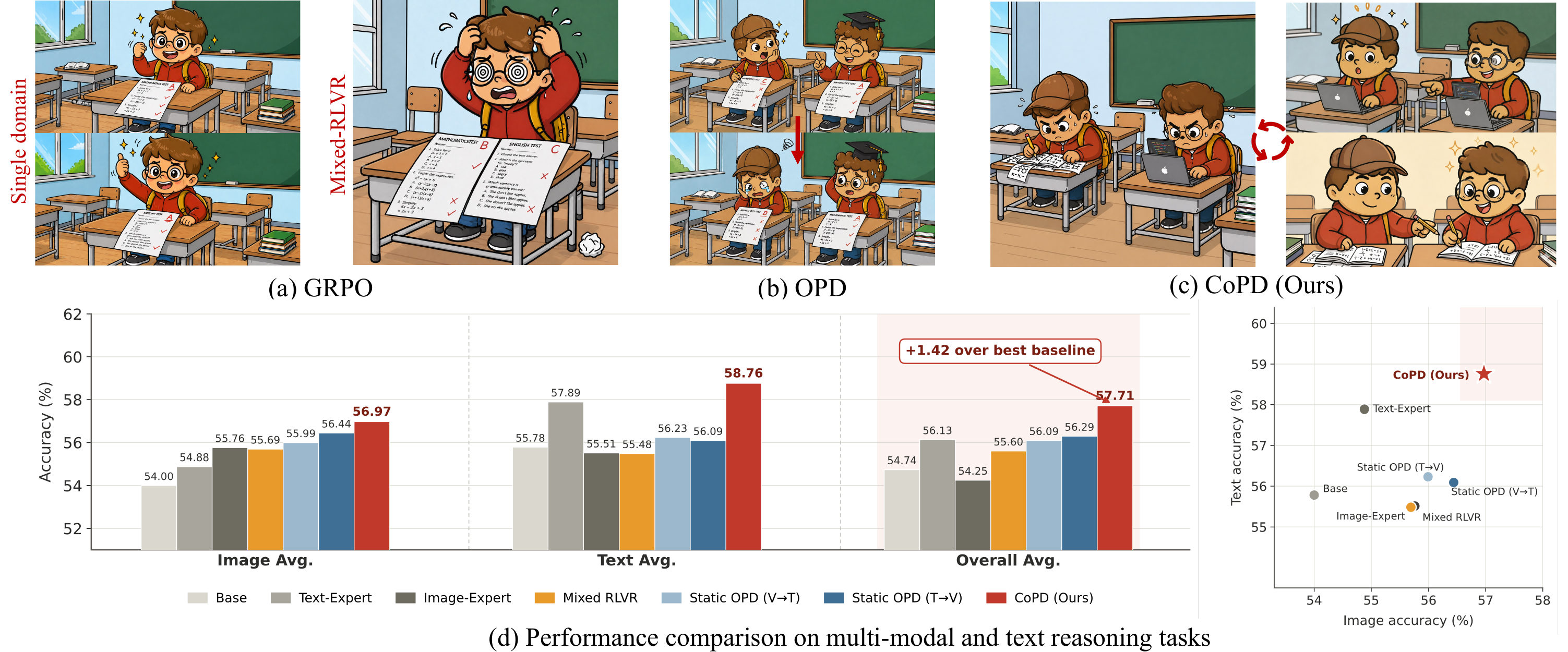}
\vspace{-0.15cm}
\caption{
CoPD addresses the limitations of mixed-data RLVR (a) and static OPD (b) by letting two branches co-evolve as teachers and students across domains (c), achieving the best overall performance.
}
\label{fig:teaser}
\end{figure*}

\input{intro}

\input{motivation-new}
\input{method}
\input{eval}
\input{related}
\input{conclusion}

\clearpage
\bibliographystyle{unsrtnat}
\bibliography{cite}

\input{appendix}

\end{document}

%% file: common.tex
\usepackage{latexsym}
\usepackage[T1]{fontenc}
\usepackage[utf8]{inputenc}
\usepackage{microtype}
\usepackage{inconsolata}
\usepackage{graphicx}
\usepackage{hyperref}       
\usepackage{url}            
\usepackage{booktabs}       
\usepackage{amsfonts}       
\usepackage{nicefrac}       
\usepackage{stackengine}
\usepackage{microtype}      
\usepackage{colortbl}
\usepackage{xcolor}
\usepackage{amsmath}
\usepackage{amssymb}
\usepackage{amsthm}
\usepackage{mathrsfs}
\usepackage{pifont}
\usepackage{MnSymbol}
\usepackage{balance}
\usepackage{enumitem}
\usepackage{listings}
\usepackage{xcolor}
\usepackage{natbib}
\usepackage{multicol}

\AtBeginDocument{%
  \providecommand\BibTeX{{%
    \normalfont B\kern-0.5em{\scshape i\kern-0.25em b}\kern-0.8em\TeX}}}

\makeatletter
\DeclareRobustCommand\onedot{\futurelet\@let@token\@onedot}
\def\@onedot{\ifx\@let@token.\else.\null\fi}

\usepackage{setspace}
\usepackage{mathtools}

\usepackage{multirow,booktabs}
\usepackage{subcaption}

\newcommand{\expect}{\mathbb{E}}

\newcommand{\owo}[1]{\textsc{OAgents}}

\definecolor{lightgreen}{RGB}{144, 238, 144} 
\definecolor{lightred}{RGB}{255, 105, 97}

\newcommand{\fakeparagraph}[1]{\vspace{1mm}\noindent\textbf{#1.}}

\newtcolorbox{promptbox}[2][Prompt]{
colback=black!5!white,
arc=5pt, 
boxrule=0.5pt,
fonttitle=\bfseries,
title=#1, 
before upper={\small}, fontupper=\fontfamily{ptm}\selectfont,
colframe=#2, 
}
\definecolor{ogreen}{RGB}{34, 139, 34}

%% file: abstract.tex
\abstract{
RLVR and OPD have become standard paradigms for post-training. We provide a unified analysis of these two paradigms in consolidating multiple expert capabilities into a single model, identifying capability loss in different ways: mixed RLVR suffers from inter-capability divergence cost, while the pipeline of first training experts and then performing OPD, though avoiding divergence, fails to fully absorb teacher capabilities due to large behavioral pattern gaps between teacher and student. We propose \textbf{Co-Evolving Policy Distillation} (\textbf{CoPD}), which encourages parallel training of experts and introduces OPD during each expert's ongoing RLVR training rather than after complete expert training, with experts serving as mutual teachers (making OPD bidirectional) to co-evolve. This enables more consistent behavioral patterns among experts while maintaining sufficient complementary knowledge throughout. Experiments validate that CoPD achieves all-in-one integration of text, image, and video reasoning capabilities, significantly outperforming strong baselines such as mixed RLVR and MOPD, and even surpassing domain-specific experts. The model parallel training pattern offered by CoPD may inspire a novel training scaling paradigm.
}

%% file: intro.tex
\section{Introduction}
Reinforcement learning with verifiable rewards (RLVR)~\cite{grpo,liu2025understandingr1zeroliketrainingcritical,yu2025dapoopensourcellmreinforcement,zheng2025groupsequencepolicyoptimization} has emerged as the dominant post-training paradigm for large models, driving rapid progress on visual reasoning~\cite{huang2026visionr1incentivizingreasoningcapability,yao2025r1sharevlincentivizingreasoningcapability}, text reasoning~\cite{deepseekr1,yang2025qwen3technicalreport}, video understanding~\cite{feng2025videor1,wang2025timer1posttraininglargevision}, and other domains~\cite{kimiteam2026kimik25visualagentic,glm5team2026glm5vibecodingagentic}. However, training a single model on mixed-capability data with RLVR often exhibits a capability trade-off~\cite{shukor2025scalinglawsoptimaldata}, where gains in one capability come at the expense of another, as illustrated in Figure~\ref{fig:teaser}(a). We refer to this phenomenon as \textit{capability divergence}: distinct capabilities favor different optimization directions, making it difficult for a single training run to advance all of them simultaneously. Since no single RLVR run can master all capabilities at once, the field has converged on a two-stage pipeline: separate copies of the base model are first trained with RLVR on each capability to obtain specialized experts, and these experts are then consolidated into a unified policy model through on-policy distillation (OPD)~\cite{lu2025onpolicy-opd}, also known as Multi-teacher OPD (MOPD) when multiple experts serve as teachers simultaneously~\cite{mimo2025flash-mopd,deepseekai2026deepseekv4}. By separating capability-specific training from cross-capability consolidation, OPD avoids the gradient conflicts caused by capability divergence
, which has made OPD a widely adopted approach for capability consolidation.

However, the way OPD is currently applied leaves an important issue unaddressed. In current practice, each expert is first trained until convergence on its own capability data, and then used as a fixed teacher for distillation into a separate student model, as illustrated in Figure~\ref{fig:teaser}(b). We argue that by the time distillation begins, the teacher has drifted too far from the student for its supervision to be effectively absorbed. To examine this, we measure the behavioral distance between teacher and student via \textit{top-$k$ token overlap} along on-policy trajectories. Our pilot study (\S\ref{sec:motivation}) shows that effective distillation requires teacher and student to remain behaviorally close. When they are too far apart, the teacher's supervision becomes difficult for the student to absorb, which is precisely where the static OPD pipeline falls short. We therefore argue that teacher and student should \textit{co-evolve}, with their behavioral distance kept within the absorbable range throughout training, rather than being set up as a static expert and a passive learner only at the moment of distillation.

Building on this insight, we propose Co-Evolving Policy Distillation (CoPD), which unifies capability exploration and consolidation into a single co-evolving process. CoPD maintains parallel training branches, each associated with a different capability. Instead of freezing each branch as a static teacher, CoPD interleaves branch-specific RLVR with cross-branch mutual distillation throughout training, so that each branch continuously deepens its own expertise while serving as an evolving teacher for the other.
Concretely, training proceeds in alternating phases, as illustrated in Figure~\ref{fig:teaser}(c). In the \textit{RLVR phase}, each branch independently optimizes on its own capability data, deepening branch-specific expertise and \textit{opening up} the behavioral distance between branches, so that subsequent distillation has substantive new knowledge to transfer. In the \textit{mutual OPD phase}, each branch generates rollouts on the other branch's data and receives token-level supervision from the other, \textit{closing} the behavioral distance and absorbing what the other branch has newly learned with high efficiency, since the two branches remain within each other's effective range. 
The two phases are tightly complementary: RLVR drives the branches apart enough to make distillation informative, while interleaving keeps them close enough for distillation to be absorbable. Mutual OPD then exploits this window to transfer what each branch has learned to the other.
Because all branches start from a shared base and remain tightly coupled through continuous mutual distillation, their parameters do not diverge drastically, and the final unified model can be obtained through simple parameter merging.

We evaluate CoPD in both two-branch and three-branch post-training settings. The two-branch setting unifies text reasoning and multimodal reasoning, with each branch specializing in one capability and serving as an evolving teacher for the other. To demonstrate that CoPD scales beyond the pairwise case, we further evaluate a three-branch setting that jointly consolidates text reasoning, image-text reasoning, and video understanding.
CoPD consistently outperforms single-domain expert, mixed-data RLVR and static (M)OPD pipelines in both the two-branch and three-branch settings. Ablation studies confirm the necessity of bidirectional distillation over unidirectional alternatives and the advantage of continuous co-evolution over one-shot distillation.

Our contributions are as follows:
\begin{itemize}
    \item We identify a key limitation of the prevailing RLVR-then-OPD pipeline through a pilot study showing that OPD gain is inversely related to teacher--student behavioral distance, indicating that distillation suffers when the two drift too far apart. We accordingly argue that teacher and student should \textit{co-evolve} via cross-branch mutual distillation, a principle that naturally extends to multi-expert parallel training.
    \item We propose CoPD, which interleaves branch-specific RLVR with cross-branch mutual OPD throughout training. RLVR drives branch-specific exploration, while mutual OPD continuously transfers cross-domain knowledge. This design supplies sufficient new knowledge while keeping it absorbable at every step, achieving all-in-one consolidation without a separate distillation stage.
    \item Experiments show that CoPD consistently outperforms mixed RLVR, original OPD, MOPD baselines across text, image, video benchmarks. Notably, CoPD breaks the conventional ceiling that a unified student cannot surpass its domain-specific experts, turning cross-domain trade-offs into mutual gains.
\end{itemize}

%% file: motivation-new.tex
\section{Unleash the Potential of RLVR and OPD}
\label{sec:motivation}

This section motivates CoPD through three steps. We first formalize the structural costs incurred by mixed-data RLVR and the static OPD pipeline under a unified framework, showing that each paradigm sacrifices part of the optimization signal in a different way (\S\ref{sec:moti-analysis}). We then hypothesize that effective OPD requires teacher and student to share similar behavioral patterns, and use top-$k$ token overlap as a measurable indicator of this property (\S\ref{sec:moti-hypothesis}). Finally, we verify both the indicator and its implications through a pilot study, which directly motivates the design of CoPD (\S\ref{sec:moti-pilot}).

\subsection{A Unified View of Existing Paradigms}
\label{sec:moti-analysis}

We consider the problem of consolidating two capabilities, represented by datasets $D_1$ and $D_2$, into a single unified policy. Let $X(D_1, D_2)$ denote the total optimization signal contained in the two datasets, that is, the capability gain that would be realized if the signal from each dataset could be perfectly applied to the unified policy. We analyze each paradigm by how much of $X$ it actually delivers, and what additional cost it incurs. The resulting utility takes the schematic form
\begin{equation}
U_{\mathcal{P}} \;\approx\; a_{\mathcal{P}} \cdot X(D_1, D_2) \;+\; b_{\mathcal{P}},
\label{eq:utility-general}
\end{equation}
where $a_{\mathcal{P}} \in [0, 1]$ measures how effectively paradigm $\mathcal{P}$ converts $X$ into actual capability, and $b_{\mathcal{P}} \le 0$ captures any additional loss the paradigm incurs beyond imperfect conversion.

\paragraph{Mixed-data RLVR: capability divergence as conflict.}
When a single model is jointly optimized on $D_1 \cup D_2$, the per-step update averages the gradients induced by the two datasets. Because distinct capabilities favor different optimization directions, their gradients disagree on capability-specific dimensions, and the disagreement manifests as gradient conflict within each parameter update. The resulting utility takes the form
\begin{equation}
U_{\text{mix}} \;\approx\; X(D_1, D_2) \;-\; \Phi(D_1, D_2),
\label{eq:utility-mix}
\end{equation}
where $\Phi(D_1, D_2) > 0$ is the \textit{capability divergence cost}, the magnitude of the conflict between the two capabilities' optimization directions. Mixed-data RLVR pays this cost as the seesaw effect: gains on one capability are partially or fully canceled by interference from the other, no matter how the data mixing ratio is tuned. In Eq.~\eqref{eq:utility-general}, this corresponds to $a_{\text{mix}} = 1$ and $b_{\text{mix}} = -\Phi < 0$.

\paragraph{Static OPD pipeline: divergence avoided, but absorption is poor.}
The static pipeline avoids gradient conflict by training each expert in isolation, so each branch optimizes its own gradient without interference and the divergence cost $\Phi$ is removed. The cost is instead paid at the consolidation stage. An OPD step trains the student on its own on-policy trajectories $y \sim \pi_\theta(\cdot \mid x)$, with teacher $\pi_T$ providing token-level supervision at each visited state $y_{<t}$:
\begin{equation}
\mathcal{L}_{\text{OPD}}(\theta) = \mathbb{E}_{x, \, y \sim \pi_\theta} \left[ \frac{1}{|y|} \sum_{t=1}^{|y|} D_{\mathrm{KL}}\big(\pi_T(\cdot \mid x, y_{<t}) \,\|\, \pi_\theta(\cdot \mid x, y_{<t})\big) \right].
\label{eq:opd}
\end{equation}
We argue that the supervision in Eq.~\eqref{eq:opd} is most effective when teacher and student behave similarly along on-policy trajectories: their token-level distributions then overlap substantially, and the teacher's predictions reinforce tokens the student is already likely to produce. When experts have been trained to convergence in isolation, however, their behavior has drifted far from the shared base, and their predictions on the student's trajectories diverge sharply from the student's own. The OPD signal becomes hard to absorb under this mismatch, and only a small fraction of the optimization signal is transferred. The resulting utility takes the form:
\begin{equation}
U_{\text{static}} \;\approx\; \eta(\mathcal{O}_{\text{low}}) \cdot X(D_1, D_2),
\label{eq:utility-static}
\end{equation}
where $\eta(\mathcal{O}) \in [0, 1]$ is an absorption-efficiency function depending on teacher--student behavioral overlap $\mathcal{O}$, and $\mathcal{O}_{\text{low}}$ denotes the low overlap that arises after experts converge in isolation. In Eq.~\eqref{eq:utility-general}, this corresponds to $a_{\text{static}} = \eta(\mathcal{O}_{\text{low}})$, which is positive but small, and $b_{\text{static}} = 0$.

\paragraph{What CoPD aims to achieve.}
The two paradigms above sacrifice the optimization signal in opposite ways. Mixed-data RLVR delivers the full signal but pays a divergence cost; the static OPD pipeline removes the divergence cost but delivers only a small fraction of the signal. A paradigm that avoids both costs must satisfy two conditions simultaneously: (i) keep capability-specific optimization separated to remove the divergence cost, and (ii) maintain teacher--student overlap at a level where $\eta(\mathcal{O})$ is high so that the optimization signal can actually be absorbed. CoPD instantiates both conditions by alternating capability-specific RLVR with cross-branch mutual OPD, and its utility takes the form
\begin{equation}
U_{\text{CoPD}} \;\approx\; \eta(\mathcal{O}_{\text{mod}}) \cdot X(D_1, D_2), \quad \eta(\mathcal{O}_{\text{mod}}) \gg \eta(\mathcal{O}_{\text{low}}),
\label{eq:utility-copd}
\end{equation}
where $\mathcal{O}_{\text{mod}}$ is the moderate overlap actively maintained by the alternating structure. The static pipeline and CoPD share the same functional form: both remove the divergence cost ($b = 0$) and deliver $\eta(\mathcal{O}) \cdot X$, differing only in the value of $\mathcal{O}$ at which OPD operates. The remainder of this section grounds these claims empirically: \S\ref{sec:moti-hypothesis} introduces a measurable indicator of $\mathcal{O}$, and \S\ref{sec:moti-pilot} shows that $\eta$ does increase with $\mathcal{O}$ and that the static pipeline systematically operates at low $\mathcal{O}$.

\subsection{A Behavioral Hypothesis and a Measurable Indicator}
\label{sec:moti-hypothesis}

The analysis above identifies the absorption-efficiency function $\eta(\mathcal{O})$ as the key determinant of OPD effectiveness, and reduces the difference between the static pipeline and CoPD to the value of $\mathcal{O}$ at which each operates. To make this analysis testable, we need a measurable indicator of $\mathcal{O}$. We posit the following hypothesis.

\begin{tcolorbox}[colback=gray!8, colframe=gray!50, boxrule=0.5pt, arc=2pt]
\textbf{Behavioral consistency hypothesis.} An OPD signal is more easily absorbed when teacher and student exhibit more similar behavioral patterns, since their on-policy trajectories then visit similar states and the teacher's predictions agree more closely with the tokens the student is likely to produce.
\end{tcolorbox}

We instantiate this notion of behavioral similarity through the \textit{top-$k$ token overlap} along on-policy trajectories. Given a teacher $\pi_T$ and a student $\pi_\theta$, we define
\begin{equation}
\mathcal{O}_k(\pi_\theta, \pi_T) = \mathbb{E}_{x, \, y_{<t} \sim \mu_\theta} \left[ \frac{|\mathrm{Top}_k(\pi_\theta(\cdot \mid x, y_{<t})) \cap \mathrm{Top}_k(\pi_T(\cdot \mid x, y_{<t}))|}{k} \right],
\label{eq:overlap}
\end{equation}
where $\mathrm{Top}_k(\cdot)$ denotes the set of $k$ tokens with the highest predicted probability and $\mu_\theta$ is the state visitation distribution induced by the student. Larger $\mathcal{O}_k$ indicates that, at the states the student actually visits during generation, teacher and student agree on which tokens are likely. The hypothesis above predicts that $\eta$ in Eq.~\eqref{eq:utility-static} and Eq.~\eqref{eq:utility-copd} should rise with $\mathcal{O}_k$, at least until $\mathcal{O}_k$ becomes so high that teacher and student are behaviorally indistinguishable and the supervision carries no new information.

\begin{figure*}[t]
  \centerline{\includegraphics[scale=0.34]{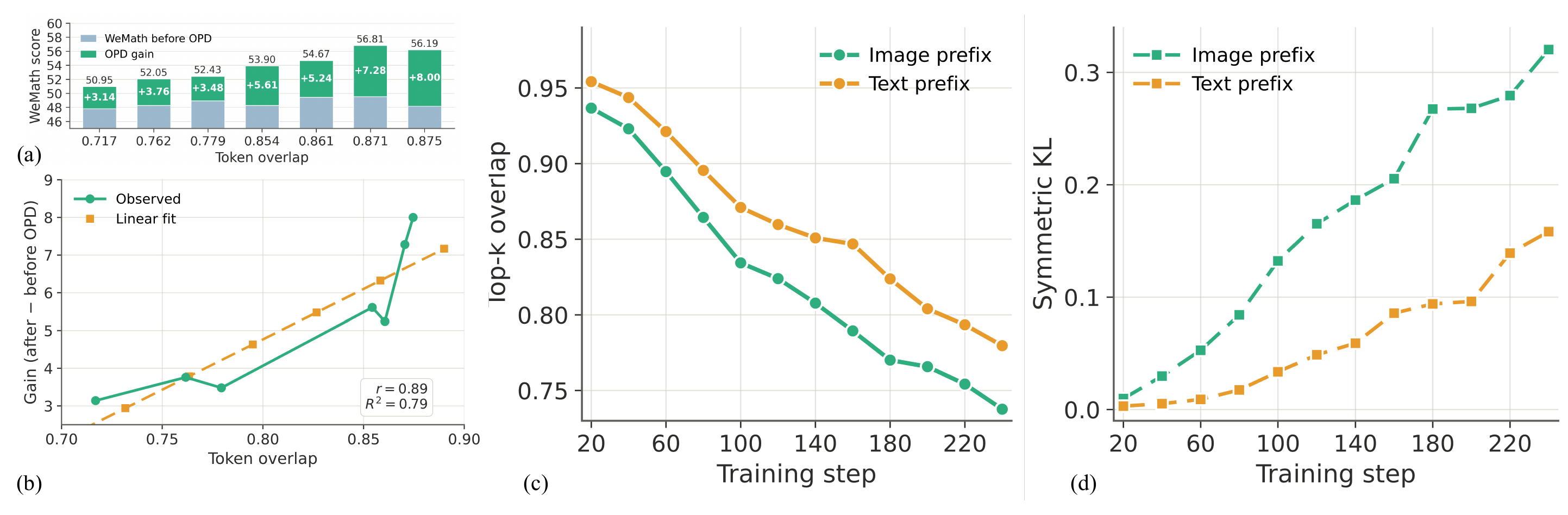}}
  \vspace{-0.1cm}
  \caption{Pilot study motivating CoPD: post-OPD gain grows with teacher--student top-$k$ overlap, while standard RLVR training pushes overlap in the opposite direction. \textbf{(a)} WeMath score before and after OPD across student checkpoints with different top-$k$ overlap to the teacher; the OPD gain (green) increases with overlap. \textbf{(b)} The post-OPD gain plotted against top-$k$ overlap, with linear fit yielding $r=0.89$ and $R^2=0.79$. \textbf{(c)} As RLVR training proceeds, top-$k$ overlap drops and symmetric KL rises, consistently across image and text branches, indicating that experts trained to convergence drift outside the effective range identified in (a) and (b).}
  \label{fig:pilot}
\end{figure*}

\subsection{Pilot Study}
\label{sec:moti-pilot}

We verify the hypothesis through two complementary experiments, summarized in Figure~\ref{fig:pilot}. The first establishes that $\eta(\mathcal{O}_k)$ does rise with $\mathcal{O}_k$ within the regime where teacher and student differ; the second establishes that the static pipeline systematically operates at a low value of $\mathcal{O}_k$.

\paragraph{Experiment 1: $\eta$ rises with teacher--student overlap.} 
We fix a teacher (an image-domain expert obtained by RLVR on $D_{\text{img}}$) and construct a series of students with varying $\mathcal{O}_k$ to the teacher by training the base checkpoint for a short period with different sampling temperatures. For each student, we run OPD on $D_{\text{img}}$ with identical training configurations and a fixed budget, and report the WeMath score before and after distillation, with $k = 10$ in $\mathcal{O}_k$.

As shown in Figure~\ref{fig:pilot}(a) and (b), the post-OPD gain increases monotonically with $\mathcal{O}_k$ across the sampled overlap range, exhibiting a strong linear correlation of $r = 0.89$ and $R^2 = 0.79$. This directly supports the behavioral consistency hypothesis: even with the teacher fixed, more behaviorally aligned students absorb the OPD signal more effectively, that is, $\eta$ in Eq.~\eqref{eq:utility-static} grows with $\mathcal{O}_k$. We do not observe the predicted decline at $\mathcal{O}_k \to 1$ in Figure~\ref{fig:pilot}(b), as temperature variation cannot drive the student all the way to teacher-identical behavior, but the boundary condition is straightforward: when teacher and student become indistinguishable, $D_{\mathrm{KL}}(\pi_T \| \pi_\theta) \to 0$ in Eq.~\eqref{eq:opd} and OPD gain necessarily collapses to zero. This boundary has its own design implication: simply running OPD continuously is not enough, since teacher and student would converge in behavior and the supervisory signal would lose informativeness. An effective paradigm must therefore actively maintain a non-trivial behavioral gap between teacher and student, in addition to keeping the gap from growing too wide.

\paragraph{Experiment 2: standard RLVR drives behavioral overlap into the low-$\eta$ regime.}
We then ask whether the static pipeline keeps OPD within the high-$\eta$ regime. We train two independent experts via RLVR, one on text-reasoning data $D_{\text{txt}}$ and one on image-reasoning data $D_{\text{img}}$, both initialized from a shared base $\pi_0$, and at fixed intervals measure each expert's top-$k$ overlap and symmetric KL with $\pi_0$ along on-policy rollouts. Treating $\pi_0$ as a proxy for the eventual student in two-stage distillation, these quantities track how teacher--student behavioral distance evolves during independent RLVR.

As shown in Figure~\ref{fig:pilot}(c)(d), $\mathcal{O}_k$ with the shared base drops monotonically as RLVR proceeds, while symmetric KL rises by an order of magnitude over the same span; the trend is consistent across both branches. This is the mirror image of the $\mathcal{O}_k \to 1$ regime discussed in Experiment 1: there, vanishing $D_{\mathrm{KL}}(\pi_T \| \pi_\theta)$ leaves the teacher with nothing new to convey; here, the growing $D_{\mathrm{KL}}$ means the teacher carries substantially more capability-specific knowledge than the student possesses, but their behavioral overlap is too low for that knowledge to be absorbed during distillation. By the time experts converge, they have moved far enough from the shared base that their overlap is well below the levels at which $\eta$ was shown to be highest in Experiment 1. This is the value of $\mathcal{O}_{\text{low}}$ in Eq.~\eqref{eq:utility-static}, and it confirms that the static pipeline systematically operates in the low-$\eta$ regime.

\paragraph{Implications for method design.}
The two experiments together expose a structural inconsistency in the static pipeline and point to a more nuanced requirement for $\mathcal{O}_k$. Experiment 1 shows that $\eta(\mathcal{O}_k)$ rises with overlap when teacher and student differ, but necessarily collapses when they become indistinguishable; Experiment 2 shows that training experts in isolation drives overlap toward the low end, where $\eta$ is also small. To raise $\eta$ from $\eta(\mathcal{O}_{\text{low}})$ to $\eta(\mathcal{O}_{\text{mod}})$ as Eq.~\eqref{eq:utility-copd} envisions, an effective paradigm must therefore satisfy three coupled requirements: (i) distillation must occur \emph{during} expert training rather than after it, so that $\mathcal{O}_k$ does not have time to drift toward the low-$\eta$ regime; (ii) the teacher must continue to evolve as the student does, so that their behavioral overlap is actively maintained rather than left to drift; and (iii) capability-specific training must continue to push the two sides apart, so that the supervision retains information the student does not already possess. The next section instantiates these three requirements as CoPD: cross-branch mutual OPD addresses (i) and (ii) by making each branch a continuously updated teacher for the other, while alternating with branch-specific RLVR addresses (iii) by periodically opening up the behavioral gap that subsequent OPD will then close.

%% file: method.tex
\section{Co-Evolving Policy Distillation}
\label{sec:method}
\begin{figure*}[t]
  \centerline{\includegraphics[scale=0.36]{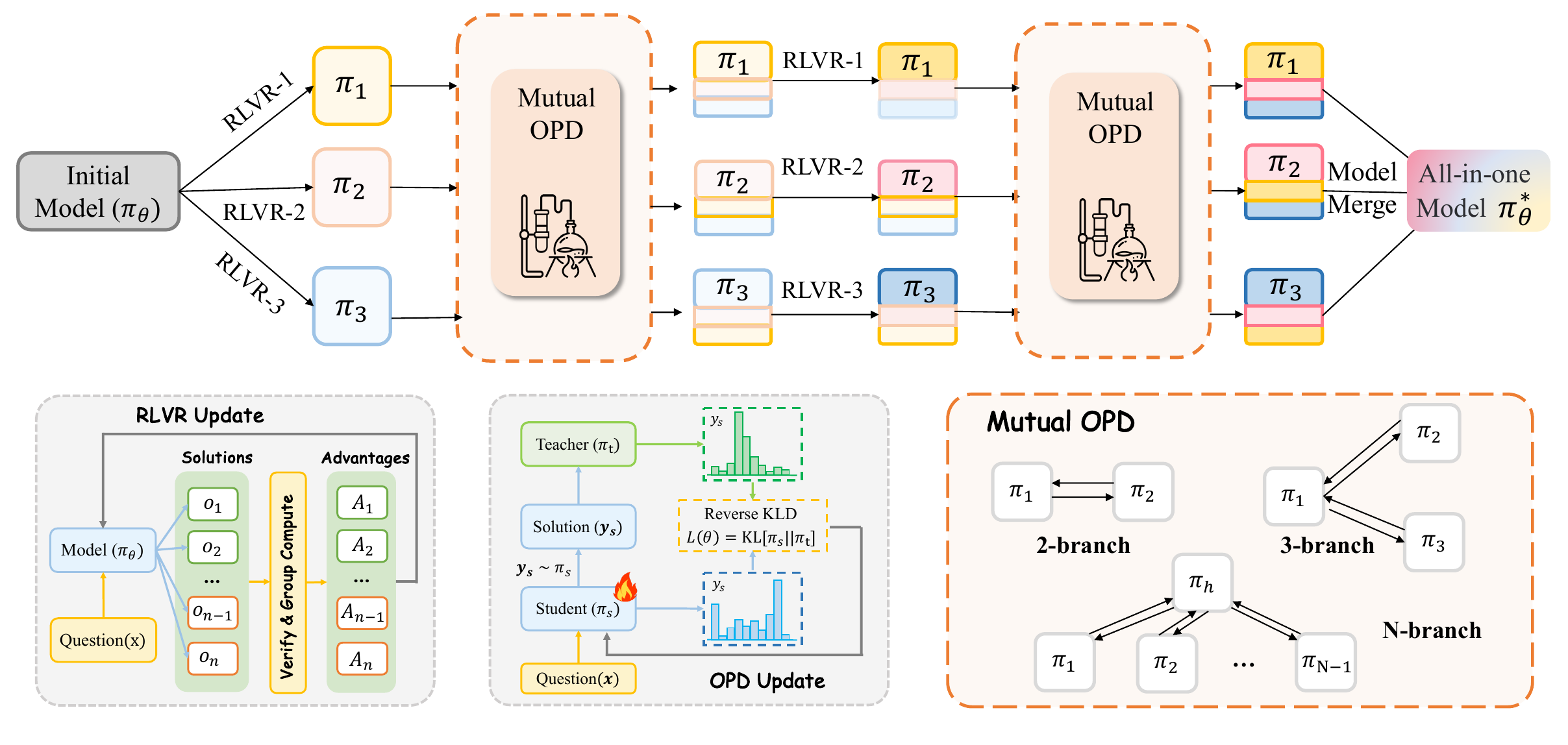}}
  \vspace{-0.1cm}
  \caption{An overview of our \method method.}
  \vspace{-0.3cm}
  \label{fig:method}
\end{figure*} 
We propose \textbf{\fullmethod (\method)}, a framework that supports multiple expert branches to co-evolve, enabling each branch to explore its own capability through RLVR and to absorb knowledge from the others through mutual on-policy distillation at appropriate intervals. We illustrate the framework with two branches. Given a base model $\pi_0$, \method initializes two parallel learning branches, $\pi_{\theta_j}$ and $\pi_{\theta_k}$, with $\theta_j^{(0)} = \theta_k^{(0)} = \theta_0$, associated with capability datasets $\mathcal{D}_j$ and $\mathcal{D}_k$, respectively. Training alternates between two phases. In the \textit{branch-specific RLVR} phase, each branch performs reinforcement learning on its own capability data, deepening branch-specific expertise. In the \textit{mutual OPD} phase, each branch generates on-policy rollouts and receives token-level supervision from the other branch, transferring newly acquired knowledge across capabilities. Through this alternating process, the branches maintain sufficient knowledge divergence to supply each other with informative supervision, while their behavioral distance remains close enough for that supervision to be effectively absorbed.

\subsection{Branch-Specific RLVR}
\label{sec:rlvr}
During the RLVR phase, each branch $k$ independently performs GRPO on the data corresponding to capability $k$ in order to deepen branch-specific expertise. For a prompt $x \sim \mathcal{D}_k$, branch $k$ samples a group of rollouts $\{y_i^{(k)}\}_{i=1}^{G}$ from its current policy $\pi_{\theta_k}$, which are evaluated by the capability-specific verifiable reward function $r_k$. The RLVR objective for branch $k$ is defined as
\begin{equation}
\label{eq:rlvr}
\mathcal{L}_{\mathrm{RLVR}}^{(k)}(\theta_k)
=
\mathbb{E}_{x \sim \mathcal{D}_k}
\left[
\frac{1}{G}\sum_{i=1}^{G}\frac{1}{|y_i|}
\sum_{t=1}^{|y_i|}
\min\!\left(
\rho_{i,t}^{(k)} \hat{A}_i^{\mathrm{RL}},
\;
\mathrm{clip}\!\left(\rho_{i,t}^{(k)}, 1-\epsilon, 1+\epsilon\right)\hat{A}_i^{\mathrm{RL}}
\right)
\right],
\end{equation}
where $\hat{A}_i^{\mathrm{RL}}$ and $\rho_{i,t}^{(k)}$ follow the standard GRPO formulation. 
This phase gradually drives the two branches toward differentiated capability frontiers, opening up a knowledge gap between them that the subsequent mutual distillation can exploit for cross-capability transfer.

\subsection{Mutual OPD}
\label{sec:opd}
In the mutual distillation phase, each branch generates on-policy rollouts on the other branch's data and receives token-level supervision from that branch, transferring newly acquired capability-specific knowledge across branches. Because both branches start from the same base and are periodically aligned through distillation, their policies remain behaviorally close throughout training, so the supervision falls on states each branch is familiar with and can be effectively absorbed.

Concretely, branch $k$ samples prompts from the other branch's dataset $x' \sim \mathcal{D}_j$ and generates rollouts from $\pi_{\theta_k}(\cdot \mid x')$. Branch $j$ then evaluates these rollouts at the token level. At each position $t$, the teacher signal is defined as:
\begin{equation}
\label{eq:delta}
\delta_{i,t}^{(k \leftarrow j)}
=
\log \pi_{\theta_j}(y_{i,t}^{(k)} \mid x', y_{i,<t}^{(k)})
-
\log \pi_{\theta_k}(y_{i,t}^{(k)} \mid x', y_{i,<t}^{(k)}),
\end{equation}
and the token-level advantage for the cross-branch update is $\hat{A}_{i,t}^{(k)} = \beta_k \, \delta_{i,t}^{(k \leftarrow j)}$, where $\beta_k$ balancing the relative contribution of cross-branch distillation. 
Crucially, branch-specific RLVR does not pause during mutual OPD; instead, the two objectives are interleaved to update the model. 
Since the same procedure applies symmetrically, both branches alternate between the role of teacher and student at every step, continuously exchanging newly acquired knowledge.

\subsection{Alternating Training Procedure}

\method organizes the two phases described above into $N$ alternating cycles. In each cycle $n = 1, \ldots, N$, all branches execute the following two phases in sequence, which we formulate as:

\paragraph{Phase I: Branch-specific RLVR.}
Each branch $k$ performs $S_{\mathrm{RL}}$ steps of GRPO on its own capability data, as described in Section~\ref{sec:rlvr}:
\begin{equation}
\theta_k^{(n,\mathrm{I})}
=
\texttt{RLVR}\!\left(\theta_k^{(n-1)};\, \mathcal{D}_k,\, r_k,\, S_{\mathrm{RL}}\right).
\end{equation}

\paragraph{Phase II: Mutual OPD.}
Each branch $k$ then performs $S_{\mathrm{OPD}}$ steps of mutual distillation as described in Section~\ref{sec:opd}:
\begin{equation}
\theta_k^{(n)}
=
\texttt{OPD}\!\left(\theta_k^{(n,\mathrm{I})};\, \mathcal{D}_j,\, \pi_{\theta_j},\, \, S_{\mathrm{OPD}}\right).
\end{equation}

The hyperparameters $S_{\mathrm{RL}}$ and $S_{\mathrm{OPD}}$ determine the rhythm between the two phases. Larger $S_{\mathrm{RL}}$ allows branches to accumulate more differentiated capability-specific discoveries during Phase~I, providing richer signals for subsequent distillation. Larger $S_{\mathrm{OPD}}$ leads to more thorough knowledge transfer during Phase~II. Their balance determines the overall training dynamics of \method, which we analyze in the ablation experiments. Because all branches start from the same base model and remain tightly coupled through continuous mutual distillation, their parameters do not diverge drastically. We exploit this by applying simple parameter merging across branches to obtain the final unified model. Notably, even without merging, each individual branch already achieves performance comparable to domain-specific experts, as shown in our ablation study (Section~\ref{sec:analysis_exps}). The full procedure is summarized in Algorithm~\ref{alg:method}.

While we have presented \method with two branches for clarity, the framework scales naturally to $K > 2$ branches. To avoid full pairwise distillation, we adopt a hub-and-spoke topology, where one branch serves as a shared hub and exchanges mutual OPD with each spoke branch. In our three-branch setting (Section~\ref{sec:main_exps}), the text reasoning branch acts as the hub, since the image and video branches are both vision-language models whose reasoning capabilities are largely grounded in the underlying LLM.

\begin{algorithm}[t]
\caption{\method: \fullmethod}
\label{alg:method}
\begin{algorithmic}[1]
\Require Base model $\pi_{\theta_0}$, $K$ capability datasets $\{\mathcal{D}_k\}_{k=0}^{K-1}$, reward functions $\{r_k\}_{k=0}^{K-1}$, total cycles $N$, RLVR steps $S_{\mathrm{RL}}$, OPD steps $S_{\mathrm{OPD}}$
\State Initialize $K$ branches: $\theta_k \gets \theta_0$ for all $k$
\For{$n = 1, \dots, N$}
    \State \textcolor{gray}{\textit{// Phase I: Branch-specific RLVR (all branches in parallel)}}
    \For{branch $k = 0, \dots, K-1$ \textbf{in parallel}}
        \State Optimize $\theta_k$ on $\mathcal{D}_k$ with GRPO (Eq.~\ref{eq:rlvr}) for $S_{\mathrm{RL}}$ steps
    \EndFor
    \State \textcolor{gray}{\textit{// Phase II: Mutual OPD (all branches in parallel)}}
    \For{branch $k = 0, \dots, K-1$ \textbf{in parallel}}
        \For{$s = 1, \dots, S_{\mathrm{OPD}}$}
            \State \textbf{Native:} generate rollouts on $\mathcal{D}_k$, update with GRPO using $r_k$
            \State \textbf{Cross-branch:} for each $j \neq k$, generate rollouts on $\mathcal{D}_j$ from $\pi_{\theta_k}$; compute teacher signal $\delta^{(k \leftarrow j)}$ from $\pi_{{\theta}_j}$ (Eq.~\ref{eq:delta}); set $\hat{A}^{(k)} = \beta_k \, \delta^{(k \leftarrow j)}$
            \State Combine native and cross-branch batches; update $\theta_k$
        \EndFor
    \EndFor
\EndFor
\State \textcolor{gray}{\textit{// Merge co-evolved branches into a unified model}}
\State $\theta^* \gets \texttt{Merge}(\theta_0, \theta_1, \dots, \theta_{K-1})$
\State \Return $\theta^*$
\end{algorithmic}
\end{algorithm}

%% file: eval.tex
\section{Experiments} 
\label{sec:exps}
\subsection{Experimental Setting}
\fakeparagraph{Training Data and Evaluation Benchmarks} 
We evaluate \method on its ability to co-evolve text, image, 
and video reasoning capabilities through parallel branch 
training. Our main analysis focuses on the two-branch setting 
with text and image reasoning; we additionally evaluate a 
three-branch setting that incorporates video reasoning to 
demonstrate scalability. For text reasoning, we use 
Polaris-Dataset-53K~\cite{Polaris2025}, filtered from 
DeepScaleR-Preview-Dataset~\cite{deepscaler2025} and 
AReal-boba-Data~\cite{fu2025areal} to retain high-quality mathematical 
reasoning problems. For image reasoning, we use 
MMFineReason-123K~\cite{lin2026mmfinereason}, a collection of image reasoning 
samples with verifiable answers. For video reasoning, we collect training data from OneThinker~\cite{feng2025onethinker}, VideoChat-R1~\cite{li2025videochatr1enhancingspatiotemporalperception}, and Video-R1~\cite{feng2025videor1}, and filter with Qwen3-8B-VL by removing samples with a pass rate of either 0\% or 100\%, retaining 40K samples of moderate difficulty.

\fakeparagraph{Evaluation Benchmarks} For evaluation, image reasoning is assessed on seven benchmarks. MMMU~\cite{yue2024mmmumassivemultidisciplinemultimodal} and MMMU-Pro~\cite{yue2025mmmuprorobustmultidisciplinemultimodal} cover multi-discipline college-level problems. MathVista~\cite{lu2024mathvistaevaluatingmathematicalreasoning} and MathVision~\cite{wang2024measuring} test mathematical reasoning with visual context. ZeroBench~\cite{roberts2025zerobenchimpossiblevisualbenchmark} evaluates zero-shot visual reasoning. WeMath~\cite{qiao-etal-2025-math} and MathVerse~\cite{zhang2024mathversedoesmultimodalllm} focus on math problems presented with diagrams and multi-format visual inputs. Text reasoning is assessed on five benchmarks. AIME 2024 and AIME 2025~\cite{aime} are competition-level mathematical olympiad problems. HMMT 2025~\cite{balunovic_srimatharena_2025} contains problems from the Harvard-MIT Mathematics Tournament. MATH-500~\cite{math500hendrycks2021measuringmathematicalproblemsolving} is a curated subset of the MATH benchmark, and Minerva Math~\cite{lewkowycz2022solvingquantitativereasoningproblemsminerva} covers scientific and quantitative reasoning. For all benchmarks, we report accuracy (\%) as the evaluation metric. We report Vis. Avg. and Text Avg. as the mean accuracy over the respective benchmark groups. For the three-branch setting, we additionally evaluate video reasoning on Video-Holmes~\cite{cheng2025videoholmesmllmthinklike}, MVBench~\cite{li2024mvbenchcomprehensivemultimodalvideo}, MMVU~\cite{zhao2025mmvumeasuringexpertlevelmultidiscipline}, and VideoMathQA~\cite{rasheed2025videomathqabenchmarkingmathematicalreasoning}, and report Video Avg. as the mean accuracy over these video benchmarks.

\fakeparagraph{Models and Baselines} We mainly conduct experiments on Qwen3-VL-4B-Instruct~\cite{bai2025qwen3vltechnicalreport} and compare \method against the following baselines. Text-Expert and Image-Expert are trained independently from the base model via RLVR on text reasoning data and image reasoning data, respectively. In the three-branch setting, we additionally train a Video-Expert on video reasoning data. Mixed 
RLVR combines all data into a single pool and trains one model via RLVR. Static OPD (V$\to$T / T$\to$V) follows a two-stage pipeline: two branches are first independently trained via RLVR on their respective domains, then one expert serves as a fixed teacher and transfers knowledge to the other through unidirectional OPD. We report results for both directions, where V$\to$T denotes the image expert teaching the text branch and T$\to$V denotes the reverse. For the three-branch setting, we use multi-teacher OPD (MOPD) as a stronger baseline, where all domain experts jointly distill into a single student. \method (Ours) alternates between RLVR and bidirectional mutual on-policy distillation throughout training.

\fakeparagraph{Implementation Details} We implement CoPD on top of the EasyVideoR1 framework~\cite{qin2026easyvideor1easierrlvideo}, which builds upon verl~\cite{sheng2024hybridflow} and EasyR1~\cite{zheng2025easyr1}. During training, the maximum input and output length are both set to 16,384 tokens. The learning rate is fixed at $1 \times 10^{-6}$. The rollout batch size is set to 256, and for each prompt we sample 8 rollouts at temperature 1.0. The clipping bounds are set to $\epsilon_{\text{low}} = 0.2$ and $\epsilon_{\text{high}} = 0.28$. Static OPD is built on top of two independently trained specifc experts and performs one additional stage of OPD. Mixed RLVR and CoPD use a total number of training steps equal to the sum of the two specifc experts to ensure the same data throughput. 

\subsection{Main Results}
\label{sec:main_exps}
\input{tables/main_results}
\fakeparagraph{Co-Evolution on Text and Image Reasoning}
Table~\ref{tab:two_branch_results} presents the main results on text and image reasoning, where \method achieves the best overall performance among all baselines. Mixed RLVR weakens text reasoning compared to the Text-Expert, confirming the capability divergence cost analyzed in \S\ref{sec:motivation}: jointly optimizing heterogeneous data incurs cross-domain interference that erodes individual capabilities. Static OPD avoids this interference by training experts separately. In the V$\to$T direction, both image and text reasoning improve over Mixed RLVR, partially resolving the cross-domain divergence. However, text reasoning still falls well short of the Text-Expert. Similarly, in the T$\to$V direction, although image reasoning benefits from the text expert's guidance, the Text-Expert's strong text capability (57.89) is only partially transferred, dropping to 56.09 in the distilled model. In both cases, a substantial portion of the teacher's knowledge fails to reach the student through post-hoc distillation. This is consistent with the analysis in Section~\ref{sec:motivation}: when experts have drifted far from the student, their thinking patterns diverge substantially, making the distillation signal difficult to absorb. In contrast, \method improves both image reasoning and text reasoning simultaneously, surpassing the specific experts on both sides. 

\fakeparagraph{Scaling to Co-Evolution on Text, Image, and Video Reasoning} Table~\ref{tab:three_branch_results} extends CoPD from the dual-branch setting to a three-branch setting, where text, image, and video reasoning capabilities are jointly optimized and consolidated. The results follow the same trend as Table~\ref{tab:two_branch_results}: CoPD achieves the best overall performance and improves over MOPD across major capability groups, showing that its effectiveness generalizes beyond pairwise capability transfer. Notably, MOPD underperforms the Video-Expert on video reasoning (58.32 vs.\ 58.75), confirming that static multi-teacher distillation struggles to absorb all experts' knowledge as the number of capability branches grows. Mixed RLVR achieves the highest video average among baselines, likely because video reasoning benefits from diverse data, but this comes at the cost of text reasoning (55.39), again exhibiting the capability divergence pattern observed in the two-branch setting. \method avoids this trade-off by co-evolving all three branches, consolidating their capabilities without sacrificing any individual domain.
\input{tables/main_tri_results}
\subsection{Analysis}
\label{sec:analysis_exps}

\fakeparagraph{Ablation Study} Table~\ref{tab:ablation_results} studies the contribution of each component in \method. Removing I-OPD causes text reasoning to drop from 58.76 to 57.41, while removing T-OPD causes image reasoning to drop from 56.97 to 56.48. In both cases, the overall performance degrades, confirming that mutual OPD in both directions is necessary: each branch's learning benefits from receiving the other's distillation signal. We also ablate the merge operation. When only the text branch is retained, text reasoning remains strong (58.61) but image reasoning drops to 56.26. Conversely, when only the image branch is retained, image reasoning holds at 56.78 but text reasoning falls to 57.17. Notably, even without merging, each single branch already surpasses both static OPD variants in Table~\ref{tab:two_branch_results} on overall performance (57.24 and 56.94 vs.\ 56.09 and 56.29), showing that co-evolution alone produces branches with well-rounded capabilities. Merging further consolidates their complementary strengths, yielding the best overall result.
\input{tables/ablation_results}

\begin{figure*}[t]
  \centerline{\includegraphics[scale=0.4]{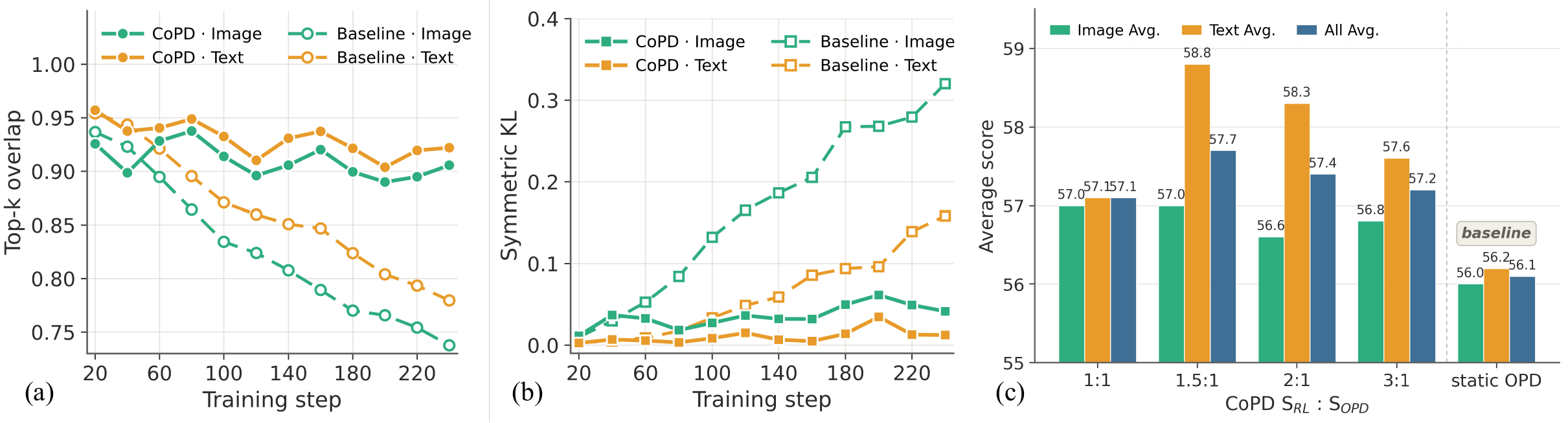}}
  \vspace{-0.1cm}
  \caption{Training dynamics and design analysis of CoPD. \textbf{(a)} Top-$k$ token overlap between the two branches throughout training: in the static-OPD baseline overlap drops monotonically, while \method maintains overlap above 0.90 via the alternation of RLVR and mutual OPD phases. \textbf{(b)} Symmetric KL between the two branches: it rises by an order of magnitude in the baseline but stays consistently low in \method. \textbf{(c)} Effect of the $S_{\mathrm{RL}}{:}S_{\mathrm{OPD}}$ ratio on final performance, with $1.5{:}1$ achieving the best overall accuracy across image, text, and overall averages.}
  \label{fig:analyse}
\end{figure*}

\fakeparagraph{Behavioral Pattern Consistency During Training}
Figures~\ref{fig:analyse} (a) and (b) track the top-$k$ token overlap and symmetric KL between the two branches throughout training. In the baseline, both metrics diverge monotonically, while symmetric KL rises by an order of magnitude. Since static OPD applies distillation after this expert training completes, it operates at the point where the two experts are furthest apart, confirming the low absorption efficiency predicted in \S\ref{sec:motivation}. In \method, top-$k$ overlap decreases during each RLVR phase but recovers during mutual OPD, staying above 0.90 throughout training. Symmetric KL remains consistently low. This confirms the core design of \method: RLVR creates divergence that makes distillation informative, and mutual OPD restores proximity that makes it easy to absorb.

\fakeparagraph{Effect of the $S_{\mathrm{RL}}/S_{\mathrm{OPD}}$ Ratio}
We further analyze the impact of the ratio between RLVR exploration steps and OPD consolidation steps in Figure~\ref{fig:analyse} (c). CoPD consistently outperforms static OPD under different $S_{\mathrm{RL}}/S_{\mathrm{OPD}}$ ratios, showing that coupling exploration with mutual distillation is more effective than post-hoc distillation. Among different ratios, $S_{\mathrm{RL}}$:$S_{\mathrm{OPD}}$ = 1.5:1 achieves the best overall performance, suggesting that sufficient branch-specific exploration is needed to create useful complementary knowledge, while overly long exploration may weaken the alignment between branches and reduce the effectiveness of subsequent distillation. This result supports the importance of balancing specialization and consolidation in CoPD.

%% file: tables/main_results.tex
\begin{table}[t]
\centering
\caption{Main results on image and text reasoning benchmarks. Mixed RLVR runs for the combined step budget of both single-domain experts. Static OPD depends on independently trained experts and performs an additional distillation stage to consolidate them. \method uses the same total steps as static OPD. Best results are in \textbf{bold} and worst results (excluding the Base Model) are marked with $^\dagger$.}
\label{tab:two_branch_results}
\setlength{\tabcolsep}{4pt}
\begin{tabular}{l ccc c cc c}
\toprule
\textbf{Benchmark} & Base & Image-Expert & Text-Expert & Mixed RLVR & \shortstack{OPD$_{V\to T}$} & \shortstack{OPD$_{T\to V}$} & \textbf{\method} \\
\midrule
\rowcolor{gray!10} \multicolumn{8}{l}{\textit{Image Reasoning}} \\
\quad MMMU          & 65.06  & 65.28  & 65.72  & 66.39  & 65.94  & 67.50  & 66.94 \\
\quad MMMU-Pro      & 53.03  & 53.89  & 53.05  & 54.28  & 54.09  & 54.71  & 55.10 \\
\quad MathVista     & 73.20  & 75.10  & 73.90  & 75.10  & 74.20  & 76.05  & 75.75 \\
\quad MathVision    & 55.07  & 55.69  & 55.82  & 56.96  & 55.82  & 56.53  & 57.88 \\
\quad ZeroBench$_\text{sub}$     & 21.41  & 21.41  & 22.01  & 21.86  & 21.41  & 21.71  & 24.40 \\
\quad WeMath        & 52.38  & 59.14  & 55.24  & 57.43  & 60.95  & 58.86  & 59.81 \\
\quad MathVerse     & 57.82  & 59.84  & 58.41  & 57.81  & 59.50  & 59.71  & 58.88 \\
\cmidrule(lr){1-8}
\rowcolor{blue!5} \quad \textbf{Avg.} & 54.00  & 55.76  & 54.88$^\dagger$  & 55.69  & 55.99  & 56.44  & \textbf{56.97} \\
\midrule
\rowcolor{gray!10} \multicolumn{8}{l}{\textit{Text Reasoning}} \\
\quad AIME 2025     & 46.88  & 47.50  & 48.33  & 44.58  & 43.54  & 45.42  & 49.58 \\
\quad AIME 2024     & 58.96  & 55.83  & 60.21  & 57.92  & 59.79  & 58.33  & 60.42 \\
\quad HMMT 2025     & 25.83  & 21.67  & 27.50  & 24.17  & 25.83  & 26.67  & 30.83 \\
\quad MATH-500      & 92.80  & 93.90  & 93.65  & 93.55  & 93.45  & 93.40  & 94.50 \\
\quad Minerva Math  & 54.41  & 58.64  & 59.74  & 57.17  & 58.55  & 56.62  & 58.46 \\
\cmidrule(lr){1-8}
\rowcolor{blue!5} \quad \textbf{Avg.} & 55.78  & 55.51  & 57.89  & 55.48$^\dagger$  & 56.23  & 56.09  & \textbf{58.76} \\
\midrule
\rowcolor{blue!5} \textbf{Overall Avg.} & 54.74  & 55.65  & 56.13  & 55.60$^\dagger$  & 56.09  & 56.29  & \textbf{57.71} \\
\bottomrule
\end{tabular}
\end{table}

%% file: tables/main_tri_results.tex
\begin{table}[t]
\centering
\caption{Main results on image, text, and video reasoning benchmarks. Mixed RLVR runs for the combined step budget of all single-domain experts. Static OPD depends on independently trained experts and performs an additional distillation stage to consolidate them. \method uses the same total steps as static OPD. Best results are in \textbf{bold} and worst results (excluding the Base Model) are marked with $^\dagger$.}
\label{tab:three_branch_results}
\setlength{\tabcolsep}{4pt}
\begin{tabular}{l cccc c c c}
\toprule
\textbf{Benchmark} & Base & Image-Expert & Text-Expert & Video-Expert & Mixed RLVR & MOPD & \textbf{\method} \\
\midrule
\rowcolor{gray!10} \multicolumn{8}{l}{\textit{Image Reasoning}} \\
\quad MMMU              & 65.06  & 65.28  & 65.72  & 66.17  & 66.11  & 66.61  & 67.78 \\
\quad MMMU-Pro          & 53.03  & 53.89  & 53.05  & 53.08  & 54.06  & 54.64  & 55.12 \\
\quad MathVista         & 73.20  & 75.10  & 73.90  & 74.85  & 75.55  & 74.75  & 76.80 \\
\quad MathVision        & 55.07  & 55.69  & 55.82  & 55.30  & 56.00  & 56.99  & 57.57 \\
\quad ZeroBench$_\text{sub}$ & 21.41  & 21.41  & 22.01  & 22.01  & 22.60  & 22.75  & 23.35 \\
\quad WeMath            & 52.38  & 59.14  & 55.24  & 53.43  & 59.05  & 59.90  & 59.90 \\
\quad MathVerse         & 57.82  & 59.84  & 58.41  & 58.16  & 59.82  & 58.93  & 59.34 \\
\cmidrule(lr){1-8}
\rowcolor{blue!5} \quad \textbf{Avg.} & 54.00  & 55.76  & 54.88  & 54.71$^\dagger$  & 56.17  & 56.37  & \textbf{57.12} \\
\midrule
\rowcolor{gray!10} \multicolumn{8}{l}{\textit{Text Reasoning}} \\
\quad AIME 2025         & 46.88  & 47.50  & 48.33  & 47.29  & 45.62  & 46.04  & 49.38 \\
\quad AIME 2024         & 58.96  & 55.83  & 60.21  & 56.46  & 54.17  & 60.83  & 60.42 \\
\quad HMMT 2025         & 25.83  & 21.67  & 27.50  & 30.83  & 25.83  & 25.00  & 31.67 \\
\quad MATH-500          & 92.80  & 93.90  & 93.65  & 93.45  & 92.95  & 93.85  & 94.35 \\
\quad Minerva Math      & 54.41  & 58.64  & 59.74  & 56.16  & 58.36  & 58.27  & 57.35 \\
\cmidrule(lr){1-8}
\rowcolor{blue!5} \quad \textbf{Avg.} & 55.78  & 55.51  & 57.89  & 56.84  & 55.39$^\dagger$  & 56.80  & \textbf{58.63} \\
\midrule
\rowcolor{gray!10} \multicolumn{8}{l}{\textit{Video Reasoning}} \\
\quad Video-Holmes       & 43.28  & 44.47  & 42.19  & 45.35  & 45.62  & 43.33  & 43.77 \\
\quad MVBench           & 68.27  & 67.82  & 65.73  & 70.86  & 67.88  & 67.71  & 69.16 \\
\quad MMVU(mc)          & 64.00  & 67.68  & 67.04  & 67.84  & 69.28  & 68.80  & 68.16 \\
\quad VideoMathQA       & 49.33  & 53.10  & 47.19  & 50.95  & 55.71  & 53.43  & 55.76 \\
\cmidrule(lr){1-8}
\rowcolor{blue!5} \quad \textbf{Avg.} & 56.22  & 58.27  & 55.54$^\dagger$  & 58.75  & \textbf{59.62}  & 58.32  & 59.21 \\
\midrule
\rowcolor{blue!5} \textbf{Overall Avg.} & 55.11  & 56.31  & 55.98$^\dagger$  & 56.39  & 56.79  & 56.99  & \textbf{58.12} \\
\bottomrule
\end{tabular}
\end{table}

%% file: tables/ablation_results.tex
\begin{table}[t]
\centering
\caption{Ablation results on image and text reasoning benchmarks with Qwen3-VL-4B. I-OPD/T-OPD denotes distillation from the image/text branch (teacher); Text-Branch Only and Image-Branch Only denote removing the merge operation and using only the text or image branch for evaluation, respectively.}
\label{tab:ablation_results}
\scriptsize
\setlength{\tabcolsep}{3.2pt}
\begin{tabular}{l cccccccc cccccc c}
\toprule
\multirow{2}{*}{Method}
& \multicolumn{8}{c}{\textit{Image Reasoning}}
& \multicolumn{6}{c}{\textit{Text Reasoning}}
& \multirow{2}{*}{All} \\
\cmidrule(lr){2-9} \cmidrule(lr){10-15}
& MMMU & M-P. & Vista & Vision & ZB$_s$ & WeM. & Verse & Avg.
& A'25 & A'24 & H'25 & M500 & Min. & Avg.
& \\
\midrule
\method
& 66.94 & 55.10 & 75.75 & 57.88 & 24.40 & 59.81 & 58.88 & \cellcolor{blue!5}56.97
& 49.58 & 60.42 & 30.83 & 94.50 & 58.46 & \cellcolor{blue!5}58.76
& \cellcolor{blue!5}57.71 \\

w/o I-OPD
& 67.39 & 55.49 & 75.30 & 57.35 & 21.71 & 61.71 & 58.50 & \cellcolor{blue!5}56.78
& 47.50 & 59.79 & 29.17 & 93.25 & 57.35 & \cellcolor{blue!5}57.41
& \cellcolor{blue!5}57.04 \\

w/o T-OPD
& 66.00 & 54.78 & 75.00 & 57.66 & 22.01 & 61.33 & 58.59 & \cellcolor{blue!5}56.48
& 49.38 & 57.71 & 28.33 & 94.10 & 59.38 & \cellcolor{blue!5}57.78
& \cellcolor{blue!5}57.02 \\

Text-Branch Only
& 65.83 & 54.25 & 76.10 & 56.86 & 24.85 & 57.62 & 58.28 & \cellcolor{blue!5}56.26
& 48.96 & 62.08 & 30.83 & 93.85 & 57.35 & \cellcolor{blue!5}58.61
& \cellcolor{blue!5}57.24 \\

Image-Branch Only
& 65.39 & 54.82 & 76.15 & 56.99 & 24.40 & 60.29 & 59.39 & \cellcolor{blue!5}56.78
& 49.38 & 59.58 & 25.00 & 94.10 & 57.81 & \cellcolor{blue!5}57.17
& \cellcolor{blue!5}56.94 \\
\bottomrule
\end{tabular}
\end{table}

%% file: related.tex
\section{Related Work}
\label{sec:related}
\subsection{RLVR for Large Models}
Reinforcement learning with verifiable rewards (RLVR) has become the dominant paradigm for eliciting reasoning capabilities in large models, replacing learned human-preference reward models with rule-based verifiable reward functions~\cite{deepseekr1,openai2024openaio1card,yu2026knowrlboostingllmreasoning,dai2025sgrpoearlyexitreinforcement}. GRPO~\cite{grpo} eliminates the need for a separate value network by estimating advantages from group-level reward statistics, making RLVR training feasible at scale, with subsequent work further improving training stability and data curation~\cite{yu2025dapoopensourcellmreinforcement,zheng2025groupsequencepolicyoptimization}. Recently, RLVR has been widely applied to enhance a variety of capabilities, including image reasoning~\cite{huang2026visionr1incentivizingreasoningcapability,yao2025r1sharevlincentivizingreasoningcapability}, video reasoning~\cite{feng2025videor1,wang2025timer1posttraininglargevision}, and coding and agentic tasks~\cite{kimiteam2026kimik25visualagentic,glm5team2026glm5vibecodingagentic}. This work explores how to accommodate multiple capabilities simultaneously within the RLVR framework.

\subsection{On-Policy Distillation}
On-policy distillation (OPD) provides teacher supervision on the student's own generated trajectories, mitigating the train-inference distribution mismatch inherent in off-policy methods~\cite{agarwal2024gkd,gu2026minillmonpolicydistillationlarge}. By offering dense token-level supervisory signals, OPD is widely used to rapidly improve a target model's capabilities with the guidance of a stronger model~\cite{lu2025onpolicy-opd, yang2025qwen3technicalreport, zhao2026opsd, hbotter2026reinforcementlearningselfdistillation_sdpo}. Recently, Multi-teacher OPD (MOPD) has been widely adopted in base model post-training, where multiple specialized experts are first trained on different capabilities and then distilled into a single policy model~\cite{mimo2025flash-mopd,deepseekai2026deepseekv4,yang2025qwen3technicalreport}. 
Through a pilot study (\S\ref{sec:motivation}), we find that this paradigm suffers from a behavioral mismatch between teacher and student that prevents the student from fully absorbing the teacher's capabilities. Our use of top-$k$ token overlap as a behavioral indicator is inspired by~\citet{li2026rethinkingonpolicydistillationlarge}, while our pilot study takes a complementary perspective by varying the student's sampling temperature to construct teacher--student pairs at controlled overlap levels. Motivated by this finding, we propose CoPD, where experts serve as teachers and students for each other, and distillation is performed during their training rather than after it. This keeps teacher and student behaviorally close while preserving the complementary knowledge gap between them, yielding more effective capability transfer.

%% file: conclusion.tex
\section{Conclusion}
This paper explores a core question: how can we better absorb the capabilities of multiple experts into a single model? We identify that both conventional approaches, mixed RLVR (i.e., directly mixing data relevant to each expert) and the traditional static OPD pipeline (first training experts, then distilling them into a policy model), suffer from non-negligible loss of expert capabilities. We propose a central idea: Co-evolving Policy Distillation, where distillation among experts should occur \textit{during} training rather than \textit{after} it, and multiple expert models should serve as teachers and students to one another, co-evolving in synergy. Each expert's corresponding branch performs RLVR on its own data to explore the capability frontier, interleaved with mutual OPD from other experts to bring their behavioral patterns closer, making them more learnable from one another. Experiments validate that CoPD achieves an all-in-one consolidation of multiple expert capabilities, even outperforming the respective expert models. The strong performance of CoPD suggests an intriguing perspective: \textit{model parallel training} may serve as a promising scaling paradigm for further broadening the boundaries of model capabilities.

This paper is the third installment of our \textit{Self-Taught RLVR} research series, which aims to investigate how LLMs can better learn from themselves and self-evolve. We explore three complementary dimensions: the first installment, RLSD~\cite{yang2026selfdistilledrlvr}, investigates the \textit{informed self}---a self augmented by privileged information that teaches the  base self; the second installment, NPO~\cite{qin2026nearfuturepolicyoptimization}, focuses on the \textit{temporal self}---a near-future self teaching its past self; this paper explores the \textit{parallel self}---parallel selves mutually teaching each other. Each of these three papers is self-contained and complete; together, they instantiate \textit{Self-Taught RLVR}.

%% file: appendix.tex
\newpage
\beginappendix
\input{preliminary}

%% file: preliminary.tex
\section{Preliminaries}
\label{sec:preliminary}

\subsection{Group Relative Policy Optimization}
Group Relative Policy Optimization (GRPO)~\cite{grpo} is a variant of Proximal Policy Optimization (PPO)~\cite{schulman2017proximalpolicyoptimizationalgorithms} tailored for large language models, which eliminates the need for a separate value network by estimating advantages directly from group-level reward statistics. Given a prompt $x$, GRPO samples a group of $G$ responses $\{y_1, y_2, \ldots, y_G\}$ from the current policy $\pi_\theta$. Each response $y_i$ receives a reward $r_i = r(x, y_i)$ from a verifiable reward function. The group-level advantage is computed by normalizing rewards within the group:
\begin{equation}
    \hat{A}_i = \frac{r_i - \operatorname{mean}(\{r_1, \ldots, r_G\})}{\operatorname{std}(\{r_1, \ldots, r_G\})}.
\end{equation}
The policy is then updated by maximizing the clipped surrogate objective with a KL regularization term against a reference policy $\pi_{\text{ref}}$:
\begin{equation}
    \mathcal{L}_{\text{GRPO}}(\theta) = \expect_{x, \{y_i\}_{i=1}^{G} \sim \pi_\theta} \left[ \frac{1}{G} \sum_{i=1}^{G} \frac{1}{|y_i|} \sum_{t=1}^{|y_i|} \left( \min\left( \rho_{i,t} \hat{A}_i,\; \text{clip}(\rho_{i,t}, 1-\epsilon, 1+\epsilon) \hat{A}_i \right) - \beta\, D_{\text{KL}}\left(\pi_\theta \| \pi_{\text{ref}}\right) \right) \right],
\end{equation}
where $\rho_{i,t} = \frac{\pi_\theta(y_{i,t} \mid x, y_{i,<t})}{\pi_{\text{old}}(y_{i,t} \mid x, y_{i,<t})}$ is the importance sampling ratio, $\epsilon$ is the clipping threshold, and $\beta$ controls the strength of the KL penalty.

\subsection{On-Policy Distillation}
On-policy distillation (OPD)~\cite{agarwal2024gkd, gu2026minillmonpolicydistillationlarge} transfers knowledge from a teacher model $\pi_T$ to a student model $\pi_\theta$ by providing token-level supervision on the student's own generated trajectories. Given a prompt $x$, the student first generates a response $y \sim \pi_\theta(\cdot \mid x)$. The teacher then evaluates this response, producing a token-level probability distribution at each position. The student is trained to minimize the KL divergence between the teacher's and its own distribution along the on-policy trajectory:
\begin{equation}
\label{eq:opd-1}
    \mathcal{L}_{\text{OPD}}(\theta) = \expect_{x,\, y \sim \pi_\theta} \left[ \frac{1}{|y|} \sum_{t=1}^{|y|} D_{\text{KL}}\left( \pi_T(\cdot \mid x, y_{<t}) \| \pi_\theta(\cdot \mid x, y_{<t}) \right) \right].
\end{equation}
Compared to off-policy distillation that trains on teacher-generated data, OPD avoids the distribution mismatch between training and inference, since the student always learns on its own trajectories~\cite{lu2025onpolicy-opd}.